\documentclass[letterpaper, 10 pt, conference]{ieeeconf}  

\IEEEoverridecommandlockouts                              

\usepackage{color}
\usepackage{amsmath}
\usepackage{amsfonts}
\usepackage{amssymb}
\usepackage{graphicx}
\PassOptionsToPackage{hyphens}{url}\usepackage{hyperref}

\usepackage{mathrsfs,amsmath}

\newcommand\revised[1]{\textcolor{black}{#1}}

\title{\LARGE \bf
  Multisensor Online Transfer Learning for 3D LiDAR-based\\
  Human Detection with a Mobile Robot
}

\author{Zhi Yan$^{1}$, Li Sun$^{2}$, Tom Duckett$^{2}$, and Nicola Bellotto$^{2}$%
  \thanks{$^{1}$LE2I-CNRS, University of Technology of Belfort-Montb\'eliard, France
    {\tt\small zhi.yan@utbm.fr}}%
  \thanks{$^{2}$Lincoln Centre for Autonomous Systems, University of Lincoln, UK
    {\tt\small \{lsun, tduckett, nbellotto\}@lincoln.ac.uk}}%
}

\begin{document}

\maketitle
\thispagestyle{empty}
\pagestyle{empty}

\begin{abstract}
Human detection and tracking is an essential task for service robots, where the combined use of multiple sensors has potential advantages that are yet to be fully exploited.
In this paper, we introduce a framework allowing a robot to learn a new 3D LiDAR-based human classifier from other sensors over time, taking advantage of a multisensor tracking system. 
The main innovation is the use of different detectors for existing sensors (i.e. RGB-D camera, 2D LiDAR) to train, online, a new 3D LiDAR-based human classifier based on a new ``trajectory probability''.
Our framework uses this probability to check whether new detection belongs to a human trajectory, estimated by different sensors and/or detectors, and to learn a human classifier in a semi-supervised fashion.
The framework has been implemented and tested on a real-world dataset collected by a mobile robot.
We present experiments illustrating that our system is able to effectively learn from different sensors and from the environment, and that the performance of the 3D~LiDAR-based human classification improves with the number of sensors/detectors used.

\end{abstract}

\section{INTRODUCTION}
\label{sec:introduction}

Human detection and tracking is an important task in service robotics, where knowledge of human motion properties such as position, velocity and direction can be used to improve the behavior of the robot, for example to improve its collision avoidance and adapt its velocity to that of the surrounding people.
%
%
Using multiple sensors to track people has advantages over a single one.
The most obvious one is that multiple sensors can often do the task with a wider field of view and thus track more people within a larger range \cite{schulz03ijrr,linder16icra}.
Another advantage is that multiple sensors providing redundant information can increase tracking accuracy and reliability \cite{kobilarov06icra,held13icra,misu14ias,koide16iros}.

\begin{figure}[t]
  \centering
  \includegraphics[height=4.8cm]{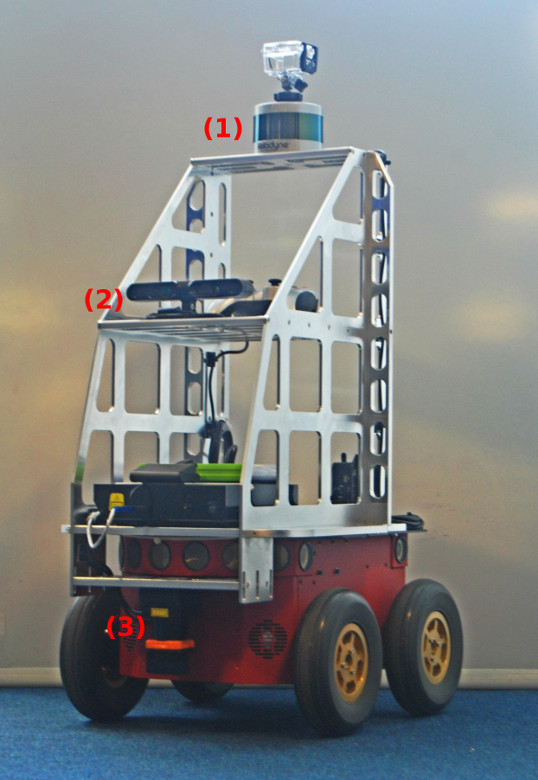}
  \hfil
  \includegraphics[height=4.8cm]{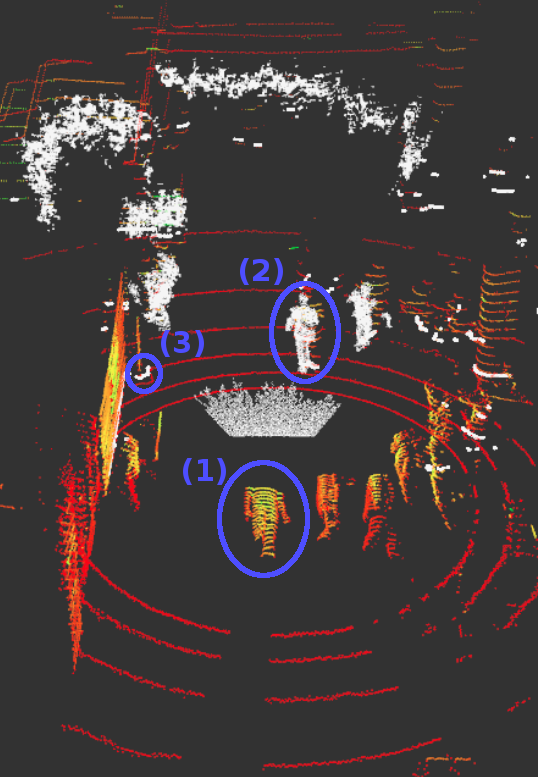}
  \caption{Left: Pioneer 3-AT mobile robot with Velodyne VLP-16 3D LiDAR (1), ASUS Xtion PRO LIVE RGB-D camera (2), and HOKUYO UTM-30LX 2D LiDAR (3).
    Right: visualized sensor data (only depth information is shown for RGB-D camera) annotated with corresponding sensor numbers.
  }
  \label{fig:flobot}
\end{figure}

Different sensors have different properties.
The 3D LiDAR in our robot platform (Fig.~\ref{fig:flobot}) has 16 scan channels, 360$^{\circ}$ horizontal and 30$^{\circ}$ vertical fields of view, and up to 100~m range.
However, this sensor provides only sparse point clouds, from which human detection can be very difficult because some useful features, such as color and texture, are missing.
2D LiDARs have obviously similar problems, with further limitations due the availability of a single scan channel and reduced field of view.
However, these sensors are also cheaper than the previous, and have been used in mobile robotics long enough to stimulate the creation of many human detection algorithms~\cite{arras07icra,bellotto2009}.
RGB-D cameras, instead, can detect humans more reliably but only within short range and limited field of view~\cite{munaro14auro,jafari14icra}.

In the literature, several algorithms already exist which can reliably detect human under particular conditions and with specific sensors (e.g. close range RGB-D detection).
Other sensors, however, are not yet so popular to benefit from good human detection software (e.g. 3D~LiDAR).
In some cases, there are simply not enough datasets with such sensors to learn robust human classifiers for many real-world applications.
In this paper, therefore, we wish to train a 3D~LiDAR-based human classifier in a semi-supervised way by learning from existing RGB-D and 2D~LiDAR-based detectors.
Although better human detectors ultimately lead to better people tracking systems, here we focus on the first part only and leave the second for future work.

Typically, data collection and training of the classifier are done offline, with obvious labor cost and potential human errors.
In the proposed transfer learning framework, instead, a 3D~LiDAR-based human classifier is trained online while exploiting spatio-temporal information from the tracking sub-system, which uses static (i.e. pre-trained) human detectors for the 2D~LiDAR and RGB-D sensors.
This framework enables a new sensor to learn from trajectories of the tracked people, each one with an associated confidence, or probability, of being human-generated, and fusing different model-based (labeled) and model-free (unlabeled) detections according to a semi-supervised learning scheme~\cite{zhu2009introduction}.
In contrast to previous approaches~\cite{Yan2017,Teichman2012}, our solution does not need any hand-labeled data, performing online learning completely from scratch. 
Besides reducing the burden of data annotation, this feature makes our system easily adaptable to the environment where the robot is deployed.


The contributions of this paper can be summarized as follows:
$1)$ we propose an online transfer learning framework for multisensor people tracking based on a new \emph{trajectory probability}, which takes into account both sensor independence (in the detection) and multisensor interaction (in the trajectory estimation);
$2)$ we present an experimental evaluation of our system for 3D~LiDAR-based human classification with a mobile robot on a real-world dataset using different sensor combinations.

The remainder of the paper is organized as follows:
Section~\ref{sec:related_work} provides an overview of relevant literature in human detection and tracking.
Section~\ref{sec:solution_framework} presents our solution framework for online transfer learning.
Section~\ref{sec:case_study} describes the application of the proposed framework to the problem of 3D~LiDAR-based human classification.
Section~\ref{sec:evaluation} illustrates the experimental results for different sensor configurations.
Finally, Section~\ref{sec:conclusions} concludes the paper summarizing the contributions and suggesting future research work.

\section{RELATED WORK}
\label{sec:related_work}

The problem of multitarget and multisensor tracking has been extensively studied during the past few decades.
Most of present systems are based on Bayesian methods~\cite{shalom95multitarget}, which compute an estimate of the correspondence between features detected in the sensor data and the different humans to be tracked.
Regarding robotic applications, multiple sensors can be deployed in single- or multi-robot systems~\cite{yz13ijars}, while the former is the concern of this paper.

RGB/RGB-D camera plus 2D LiDAR is the most frequently used combination in the literature.
\cite{kobilarov06icra} presented two different methods for mobile robot tracking and following of a fast-moving person in an outdoor environment.
The robot was equipped with an omnidirectional camera and a 2D LiDAR.
\cite{spinello09ijrr} presented an integrated system to detect and track people and cars in outdoor scenarios, based on the information retrieved from a camera and a 2D LiDAR on an autonomous car.
\cite{linder16icra} introduced a people tracking system for mobile robots in very crowded and dynamic environments.
Their system was evaluated with a robot equipped with two RGB-D cameras, a stereo camera and two 2D LiDARs.

Our previous work with the aforementioned sensor combination includes \cite{bellotto2009} and \cite{Dondrup2015a}.
The former presented a human tracking system for mobile service robots in populated environments, while the latter extended this system to a fully integrated perception pipeline for people detection and tracking in close vicinity to the robot.
The proposed tracking system tracks people by detecting legs extracted from a 2D LiDAR and fusing this with the faces or the upper-bodies detected with a camera using a sequential implementation of the Unscented Kalman Filter (UKF).

The combination with 3D LiDAR is increasing with the development of the 3D LiDAR technology.
Taking advantage of its high accuracy, \cite{held13icra} developed an algorithm to align 3D LiDAR data with high-resolution camera images obtained from five cameras, in order to accurately track moving vehicles.
Other reported results include \cite{premebida14iros} and \cite{gonzalez15iv}, which mainly focused on pedestrian detection rather than tracking.
In addition, earlier work presented multitarget tracking with a mobile robot equipped with two 2D LiDARs, respectively located at the front and back~\cite{schulz03ijrr}.
Thus the robot can have a 360$^{\circ}$ horizontal field of view, where each scan of these two sensors covers the whole surrounding of the robot at an angular resolution of 1$^{\circ}$.

The use of machine learning algorithms for tracking has particular advantages.
The closest work to ours is \cite{Teichman2012}, where the authors proposed a semi-supervised learning approach to the problem of track classification in 3D LiDAR data, based on Expectation-Maximization (EM) algorithm.
In contrast to our approach, their learning procedure needs a small set of seed tracks and a large set of background tracks, that need to be manually or semi-manually labeled at first, whereas we do not need any hand-labeled data.

To our knowledge, no existing work in the robotics field explicitly exploits information from multisensor-based tracking to implement transfer learning between different sensors as in this paper.
Our work combines the advantages of multiple sensors with the efficiency of semi-supervised learning, and integrates them into an single online framework applied to 3D~LiDAR-based human detection.

\section{ONLINE TRANSFER LEARNING}
\label{sec:solution_framework}

\begin{figure}[t]
  \centering
  \includegraphics[width=\columnwidth]{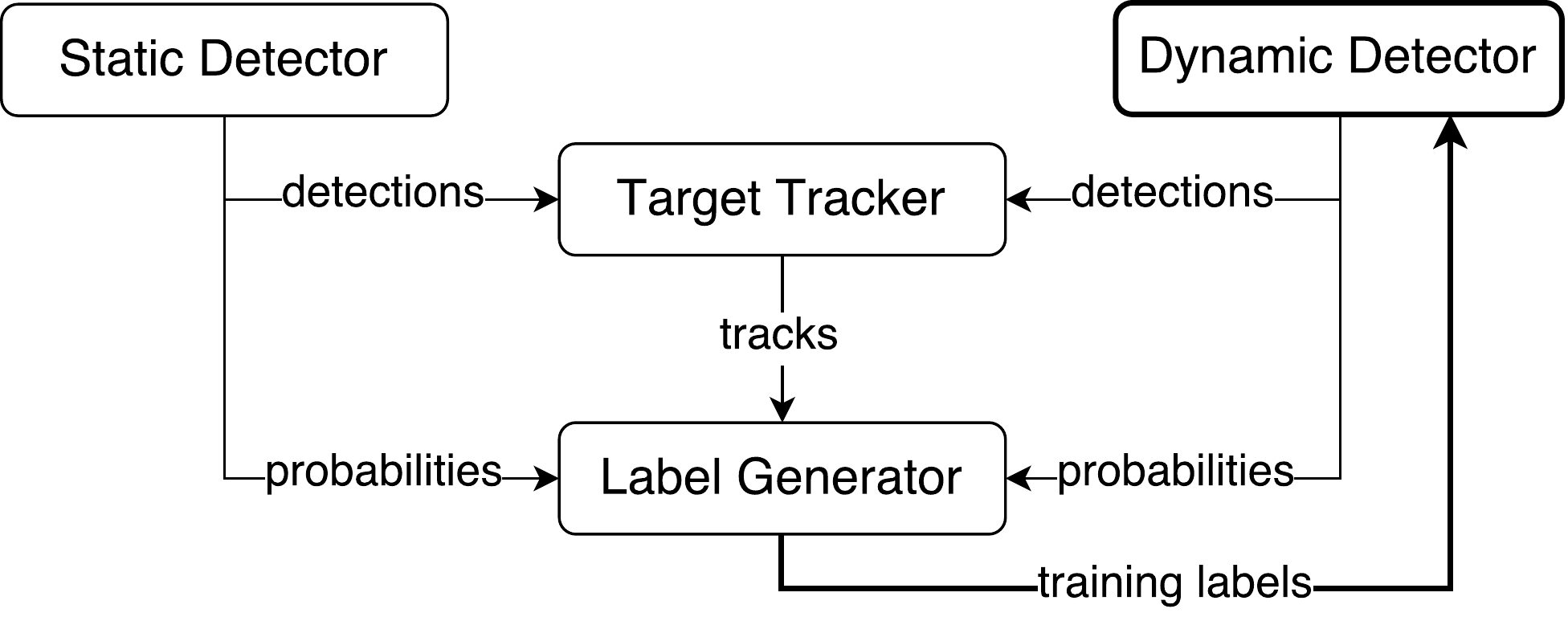}
  \caption{Block diagram of the online transfer learning framework.}
  \label{fig:framework}
\end{figure}

An overview of our solution framework for online transfer learning can be seen in Fig.~\ref{fig:framework}.
It contains four main components: static detectors denoted by $\mathcal{D}_s$, dynamic detectors $\mathcal{D}_d$, a target tracker $\mathcal{T}$ and a label generator $\mathcal{G}$.
In order to facilitate the explanation, we present each component following the sequence of an entire iteration, starting with human detection.

\subsection{Human Detection and Tracking}
\label{sec:detection_tracking}

$\mathcal{D}_s$ can detect humans with offline-trained or heuristic detectors, typically with high confidence, while $\mathcal{D}_d$ acquires this ability through the online framework.
Both detectors provide new observations for $\mathcal{T}$ and their corresponding probabilities for $\mathcal{G}$.
$\mathcal{D}_s$ provides labeled detections, while $\mathcal{D}_d$ provides both labeled and unlabeled ones.
Here we assume the initial training set is substituted, instead, by a transfer learning process between the initial $\mathcal{D}_s$ and the final $\mathcal{D}_d$.



The tracking process $\mathcal{T}$ gathers the observations, fuses them and generates human motion estimates.
Both moving and stationary targets are tracked.
For the latter, the trajectory length is supposed to be null or at least very small.
$\mathcal{T}$ associates human detections from different sensors to the same corresponding estimates, linking $\mathcal{D}_s$ and $\mathcal{D}_d$ detections and therefore making the transfer learning possible.
In order to enable this on a mobile robot with multiple sensors, $\mathcal{T}$ should: 
$a)$ be robust to sensor noise and partial occlusions;
$b)$ fuse multisensor data;
$c)$ be able to deal with multiple targets simultaneously;
$d)$ cope with noise introduced by robot motion.

\subsection{Transfer Learning}

The label generator $\mathcal{G}$ fuses the information coming from $\mathcal{D}_s$, $\mathcal{D}_d$ and $\mathcal{T}$, then generates training labels for $\mathcal{D}_d$.
A \emph{trajectory probability} is measured by $\mathcal{G}$, based on Bayes' theorem.
The idea is to measure the likelihood that a trajectory belongs to a human, which is defined as follow.
Given an objectness proposal $x_i$ and its category label $y_i$, $P(y_i|x_i,d_j)$ denote the predictive probability that sample $x_i$ is a human observed by detector $d_j \in \mathcal{D}$ ($\mathcal{D} = \mathcal{D}_s \cup \mathcal{D}_d$) at time $t$.
For a whole trajectory of detections $X_T\{x_i\}$ and its category label $Y_T$, the predictive probability of the whole trajectory $P(Y_T|X_T,\mathcal{D})$ is computed by integrating the observations of the different detectors according to the following formula:
\begin{equation}
  \label{eq:px}
  \small
  P(Y_T|X_T,\mathcal{D}) = \frac{odds_{X_{T}}}{1+odds_{X_{T}}}
\end{equation}
where
\begin{equation}
  \label{eq:odds_prod}
  \small
  odds_{X_{T}} = \prod_{i=1}^{t} \prod_{j=1}^{K} odds_{x_{i}}^j
\end{equation}
and
\begin{equation}
  \label{eq:odds}
  \small
  odds_{x_{i}}^j = \frac{P(y_i|x_i,d_j)}{1-P(y_i|x_i,d_j)}
\end{equation}
Here, $P(y_i|x_i,d_j)$ may be necessary to get through some transformations based on specific circumstances, that are illustrated in our application case (Section \ref{sec:static_detecors} and \ref{sec:dynamic_detecor}).
The above formula has been proved effective by theoretical and experimental analysis in generating grid occupancy map from different sensors \cite{burgard00icra,moravec88sensor},
since the odds method takes sensor interactions into account.
To the best of our knowledge, applying it to human trajectory analysis is new.

An intuitive approach to generate labels is to threshold the trajectory probability in Eq.~\ref{eq:px}.
Let $\sigma_t$ be a predefined threshold value, then:
\begin{equation}
  \label{eq:track_label}
  \small
  \begin{split}
    \forall x_i \in X_T & \text{~assign~} y_i = z (z \in \mathbb{Z}^+) \text{,} ~\text{get}~ X_T, Y_T = \{x_i,y_i\}_T\\
    & \text{~if~} P(Y_T|X_T,\mathcal{D}) \geq \sigma_t ~~~~
  \end{split}
\end{equation}
where $\mathbb{Z}^+$ refers to positive integer values.
Our framework uses a Batch Incremental Training (BIT) policy to learn a classifier in $\mathcal{D}_d$, because:
$a)$~the number of new generated training examples is always more than one;
$b)$~the batch of examples can be used to approximate the distribution of the whole examples in online learning~\cite{read12ida}.

For a certain interval, a new batch of training samples $[X_{new}, Y_{new}]$ can be generated:
\begin{equation}
  \label{eq:batch}
  \small
  \begin{split}
    [X_{new}, Y_{new}] = \{(x_{i}, y_{i}) ~|~ & x_{i} \in X_T \subseteq \mathbb{R}^n,\\
    & y_{i} \in Y_T \subseteq \mathbb{Z}^+, 1 \leq i \leq n_i\}_{\forall T}
  \end{split}
\end{equation}
where $X_T$ is obtained by the tracker and $Y_T$ is obtained from Eq.~\ref{eq:track_label}.
The classifier of $\mathcal{D}_d$ will be updated in the BIT procedure:
\begin{equation}
  \small
  \mathcal{D}_d^i = \text{BIT}([X_{new}, Y_{new}], \mathcal{D}_d^{i-1})) 
\end{equation}
where subscript $i$ refers to the iteration rather than time.


\subsection{Convergence of the Learning Process}

The learning procedure converges when the number of correct detections output by $\mathcal{D}_d$ reaches a steady state:
\begin{equation}
  \label{eq:halting}
  \small
  \lim_{I \to \infty}\frac{stability_I}{I} = 0
\end{equation}
where $I$ is the total number of re-training iterations.
In our system, the stability at iteration $i$ is defined as follows:
\begin{equation}
  \label{eq:stability}
  \small
  stability(I) = \sum_{i=1}^{I} \|u_i-u_{i+1} \|
\end{equation}
\begin{equation}
  \small
  \begin{aligned}
    u_i~=~&\sum_{N_i} y_i \cdot \psi (P(y|X_{valid}^i, \mathcal{D}_d^{i}))\\
    & + (1-y_{i}) \cdot \psi (1-P(y|X_{valid}^{i}, \mathcal{D}_d^{i}))
  \end{aligned}
\end{equation}
where $u_i$ is the number of correct classified examples, $X_{valid}$ is the validation set, $\psi$ is binary function via thresholding 0.5.
One can halt the learning process when the stability stops increasing, or other stopping conditions (e.g. number of iterations) are triggered.

\section{APPLICATION TO MULTISENSOR HUMAN DETECTION}
\label{sec:case_study}



In this section, we present an online transfer learning for human classification in 3D~LiDAR scans using the robot shown in Fig.~\ref{fig:flobot}.
The sensor configuration resembles the one adopted for an industrial floor washing robot developed by the EU project FLOBOT\footnote{\url{http://www.flobot.eu}} and, besides a 3D~LiDAR on the top, includes an RGB-D camera and a 2D~LiDAR mounted on the front.
We describe in the following paragraphs how to use the state-of-the-art detectors to train a 3D LiDAR-based human detector online, instead of training it offline using manually-labeled samples.

\begin{figure}[t]
  \centering
  \includegraphics[width=\columnwidth]{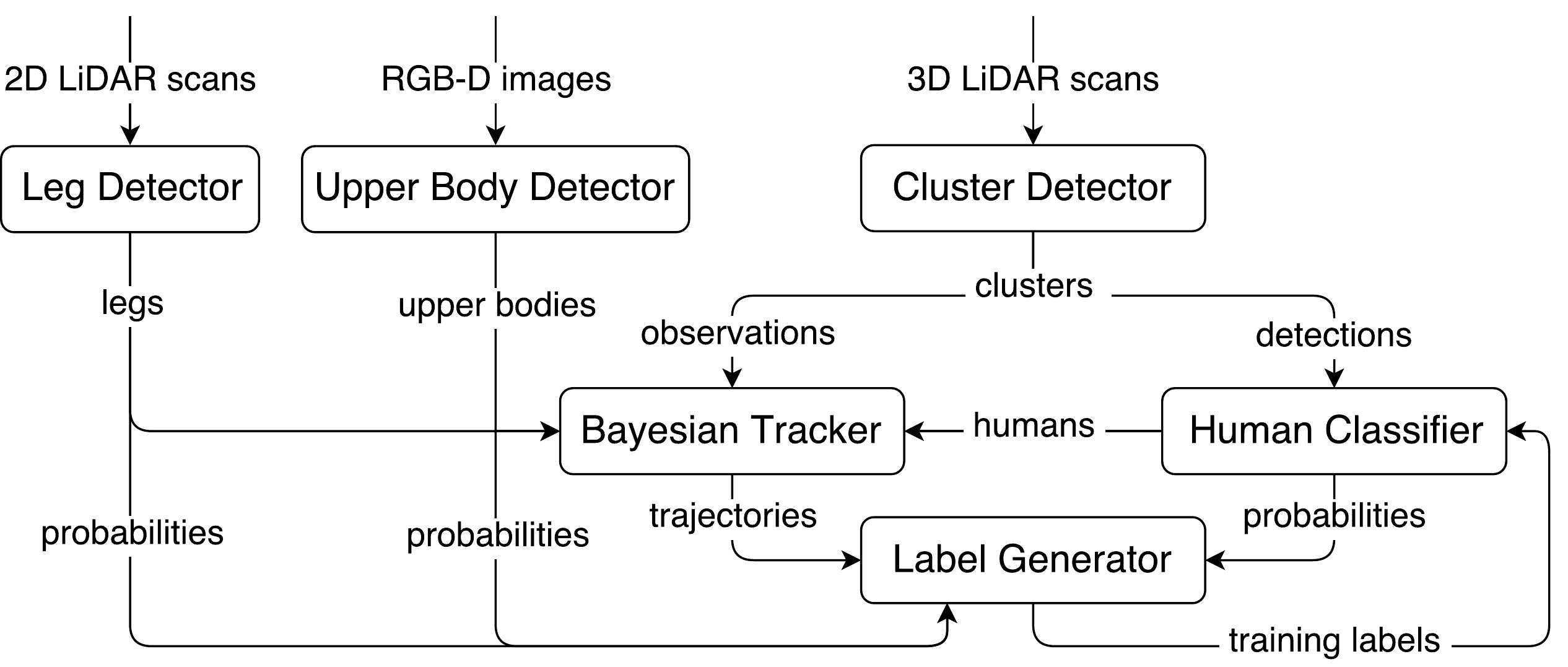}
  \caption{Process details of the online transfer learning for human detection.}
  \label{fig:case}
\end{figure}

The detailed block diagram of our implementation can be seen in Fig.~\ref{fig:case}.
At each iteration, 3D LiDAR scans are first segmented into point clusters.
The 2D position and velocity of these clusters are estimated in real-time by a multitarget tracking system, which outputs the trajectories of all the clusters.
At the same time, a classifier is trained to classify the clusters as human or not, assigning a normalized confidence value to each of them.
This confidence is the predictive probability $P(y_i|x_i, d_j)$ for the 3D~LiDAR-based detector, which is needed for the calculation of the trajectory probability in Eq.\ref{eq:px}-\ref{eq:odds}.
The classifier is initialized and retrained online.
The trajectories and the probabilities are sent to a label generator, which generates the training labels for the next iteration.

\begin{figure}[t]
  \centering
  \includegraphics[width=\columnwidth]{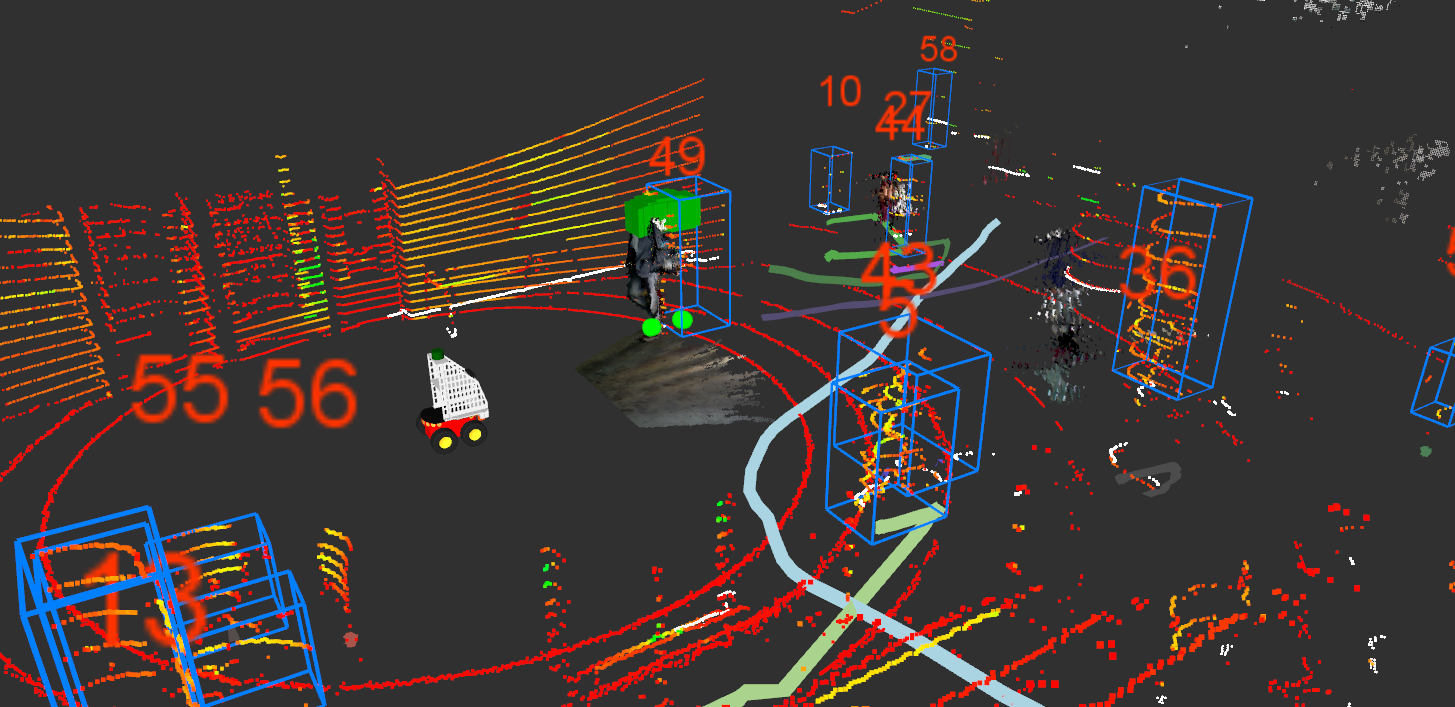}
  \caption{A screenshot of our multisensor-based detection and tracking system in action.
    The sparse colored dots represent the laser beams with reflected intensity from the 3D~LiDAR.
    The white dots indicate the laser beams from the 2D~LiDAR.
    The colored point clouds are RGB images projected on depth data of the RGB-D camera.
    The robot is at the center of the 3D LiDAR beam rings.
    The numbers are the tracking IDs and the colored lines represent the people trajectories generated by the tracker.
    For example, the person with tracking ID~49 has been detected by the RGB-D upper-body detector (green cube), the 2D~LiDAR leg detector (green circle), and the 3D~LiDAR cluster detector (blue bounding box).}
  \label{fig:screenshot}
\end{figure}

The upper-body detector~\cite{jafari14icra} and the leg detector~\cite{arras07icra}, respectively based on the RGB-D camera and the 2D~LiDAR, are the static detectors $\mathcal{D}_s$.
Both enable human tracking by sending the position of the detections.
In addition, they provide the corresponding probabilities $P(y_i|x_i, d_j)$ (i.e. normalized detection confidence) to the label generator.
The combination of 3D-LiDAR-based cluster detector and the human classifier, instead, constitutes the dynamic detector $\mathcal{D}_d$ that we want retrain online.
For an intuitive understanding of the various detectors and their outputs, please refer to the example in Fig.~\ref{fig:screenshot}.
The following paragraphs describe each module in detail.

\subsection{Upper Body Detector and Leg Detector}
\label{sec:static_detecors}


The upper-body detector identifies upper-bodies (shoulders and head) in 2D range (depth) images, taking advantage of a pre-defined template.
The confidence of the detection is inversely proportional to the observation range.
The leg detector detects legs in 2D LiDAR scans based on 14 features, including the number of beams, circularity, radius, mean curvature, mean speed, and more.
Its detection performance is limited by moving and crowd people.
As the upper-body detector and leg detector are not probabilistic methods and for the sake of simplifying mathematical conversion, a probability of 0.5 is assigned if $x_i$ is detected as a human.

\subsection{Cluster Detector and Human Classifier}
\label{sec:dynamic_detecor}


The 3D~LiDAR-based cluster detector and the human classifier are originally from our recent work~\cite{Yan2017}, while the former has been incorporated in different problems \cite{ls18icra, ls18ral}.
As input of this module, a 3D LiDAR scan is first properly segmented into different clusters using an adaptive clustering approach.
The latter enables to use different optimal thresholds for point cloud clustering according to the scan ranges.
Then, a Support Vector Machine (SVM)-based classifier \cite{svm} with six features (a total of 61 dimensions) is trained online.
These features are selected to meet robots' requirements for real-time and online computing performance.
For more implementation details, please refer to~\cite{Yan2017}.
%

In our approach (based on LIBSVM~\cite{libsvm}), the uncalibrated error function of SVM is squashed into a logistic function (here is the sigmoid function) to get the predictive probability $P(y_i|x_i,d_j)$ (used in Eq.~\ref{eq:odds}).
To be more specific, a binary classifier (i.e. human or non-human) is trained at each iteration. 
The ratio of positive to negative training samples is set to $1:1$, and all data are scaled to $[-1, 1]$, generating probability outputs and using a Gaussian Radial Basis Function kernel~\cite{rbf}.
Since LIBSVM does not currently support incremental learning, the system stores all the training samples accumulated from the beginning and retrains the entire classifier at each new iteration.
The solution framework, however, also allows for other classifiers and learning algorithms.


\subsection{Bayesian Tracker}

People tracking is performed by a robust multisensor-multitarget Bayesian tracker, which has been widely used and described in previous works~\cite{linder16icra,bellotto2009,Yan2017,Dondrup2015a,BayesianTracking}.
The estimation consists of two steps.
In the first step, a constant velocity model is used to predict the target state at time $t$ given the previous state at ${t-1}$.
%
In the second step, if one or more new observations are available from the detectors, the predicted states are updated using a Cartesian or a Polar observation model, depending on the type of the sensor.
An efficient implementation of Nearest Neighbor data association is finally included to resolve ambiguities and assign each person the correct detections, in case more than one are simultaneously generated by the same or different sensors. As reported in~\cite{linder16icra,Dondrup2015a,BayesianTracking}, our tracker fulfills the requirements listed in Sec.~\ref{sec:detection_tracking}.
We refer the reader to these publications for further details.

\subsection{Label Generator}


The positive training labels $X^+$ are generated according to Eq.~\ref{eq:track_label}, while the negatives $X^-$ are generated based on a volume filter:
\begin{equation}
  \label{eq:volume_filter}
  \small
  \begin{aligned}
    X^-~=~&\{x_i~|~\textsc{w}_i < 0.2,~\textsc{w}_i > 1.0,\\
    & \textsc{d}_i < 0.2,~\textsc{d}_i > 1.0,~\textsc{h}_i < 0.2,~\textsc{h}_i > 2.0\}
  \end{aligned}
\end{equation}
where $\textsc{w}_i$, $\textsc{d}_i$, $\textsc{h}_i$ are the width, depth and height of a 3D cluster $x_i$.
The idea is that clusters without a pre-defined human-like volumetric model will be considered as negative samples for the next training iteration.
In our application, the dynamic classifier was trained from scratch without any manually-labeled initial sets.
As the validation set is not available, the maximum number of iterations was used as halting criteria.

\section{EVALUATION}
\label{sec:evaluation}

\subsection{Dataset}

We evaluated our framework on a real-world dataset\footnote{\url{https://lcas.lincoln.ac.uk/wp/research/data-sets-software/l-cas-multisensor-people-dataset/}} collected in an indoor public area by the robot shown in Fig.~\ref{fig:flobot}.
The robot was running the Robot Operating System (ROS)~\cite{ros} and it was manually driven with a gamepad.
Several \emph{rosbag} files were recorded, and the total length of which is about 49 minutes, including two continuous recordings of 19 and 30 minutes.
Sensor data were recorded in their original frame of reference and the coordinate transformations were handled by the ROS \emph{tf} package.
%

\subsection{Experimental Setup}

The experiments were conducted on the 19 minutes segment of continuous data, in which our binary SVM human classifier was learned online.
The classifier was retrained once every 300 new positive (human) and 300 new negative (non-human) samples, \revised{labeled by the label generator}, corresponding to one iteration.
We report the results for the first seven iterations, collecting a total of 2,100 positive and 2,100 negative.
In addition, a classifier was trained offline, using 2,100 manually labeled positive samples with an equal amount of randomly selected negative samples, to serve as a baseline for comparison.
Furthermore, we arbitrarily selected 100 scan frames from the dataset and fully annotated these (including standing and sitting people) as a test set.
This contains 1,197 human labels with varying distances from the robot between few centimeters and twenty meters.
A detection was considered a true positive if the overlap between it and the ground truth was more than~50\%.

Our framework has been fully implemented within a modular ROS architecture.
All components are ready for download\footnote{\url{https://github.com/LCAS/online_learning/tree/multisensor}} and use by other researchers.
The dataset collection and all the experiments reported in this paper were carried out on the robot embedded PC, with an Intel i7-4785T processor and 8~GB memory, using Linux Ubuntu 14.04~LTS (64-bit) and ROS Indigo.
It is worth noting that our system is fast and cost effective, since it can learn a human detector within minutes and using only inexpensive CPUs, rather than training for hours or days with expensive GPUs.

\subsection{Human Classification}

We first evaluate the performance of the 3D~LiDAR-based human classification after every online training iteration.
We compare the results for all the possible sensor combinations: 3D~LiDAR only, i.e. without any knowledge transfer \revised{but learned from trajectories only}; 3D~LiDAR with RGB-D camera; 3D~LiDAR with 2D~LiDAR; 3D~LiDAR with both RGB-D camera and 2D~LiDAR.
We measure the average precision (AP)~\cite{voc}, rather than the classification accuracy (ACC) used in other methods~\cite{Teichman2012}, because more informative.
Indeed the number of true negatives in our binary classification was far larger than the number of true positives, leading to an ACC always higher than 80\% (with probability threshold $\sigma_t = 0.5$), for each training iteration and each sensor combination.
In all the experiments, the trajectory probability threshold $\sigma_t$ of Eq.~\ref{eq:track_label} was set to $0.7$.

\begin{figure}[t]
  \centering
  \includegraphics[width=0.92\columnwidth]{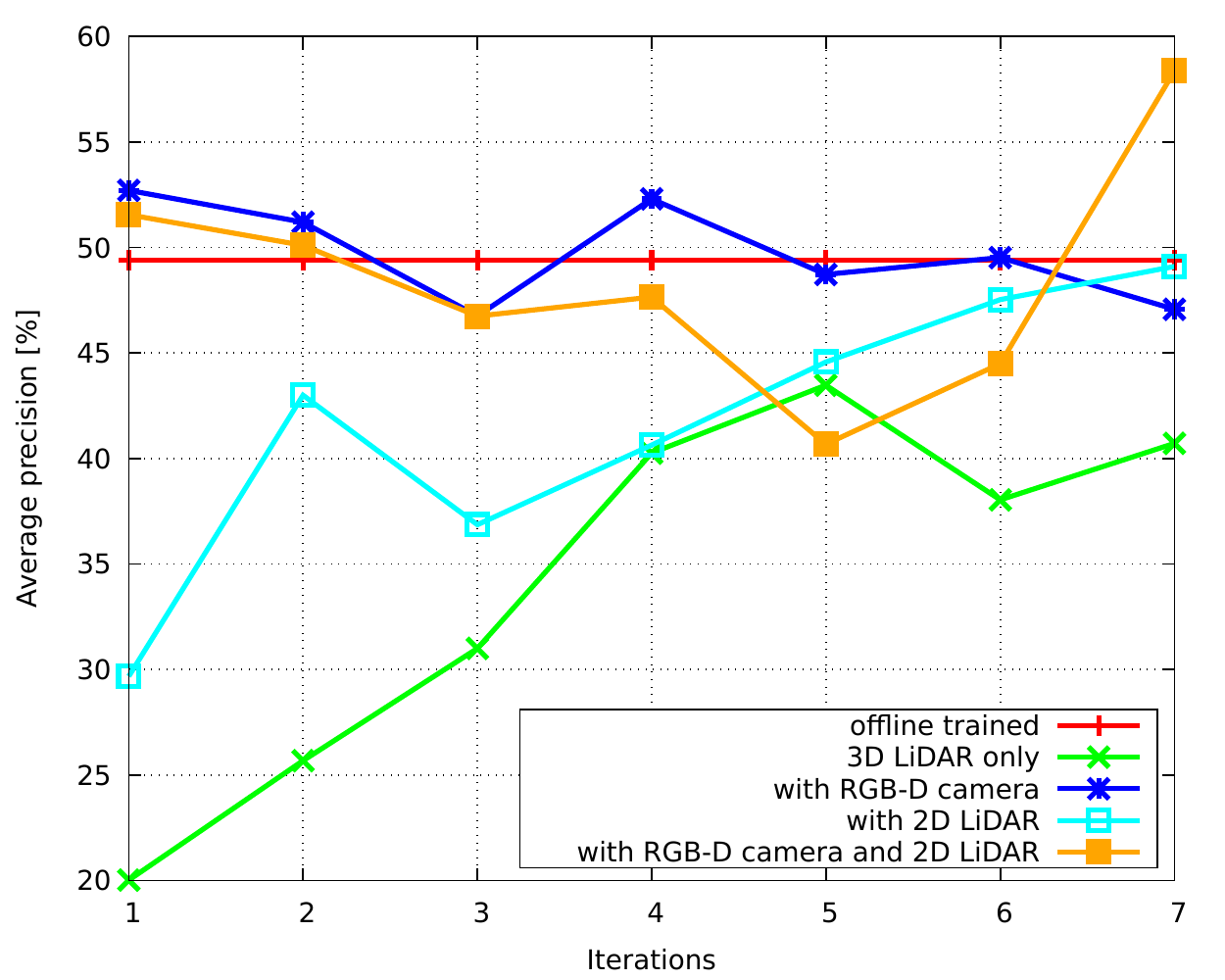}
  \caption{Comparison of different 3D~LiDAR-based human classifiers trained offline (red line) and online using different sensor combinations. Best viewed in color.}
  \label{fig:detection_results}
\end{figure}

The experimental results in Fig.~\ref{fig:detection_results}, show that the AP for ``3D~LiDAR only'' and (3D~LiDAR) ``with 2D~LiDAR'' increases with the iterations, while there are not significant changes ``with RGB-D camera''.
However, the results ``with RGB-D camera and 2D~LiDAR'' show an interesting trend, first decreasing until the 5th iteration, and then increasing well above any other combination.
This final outcome shows the advantage of our multisensor system, which can eventually improve the online transfer learning process.

\begin{figure}[t]
  \centering
  \includegraphics[width=0.92\columnwidth]{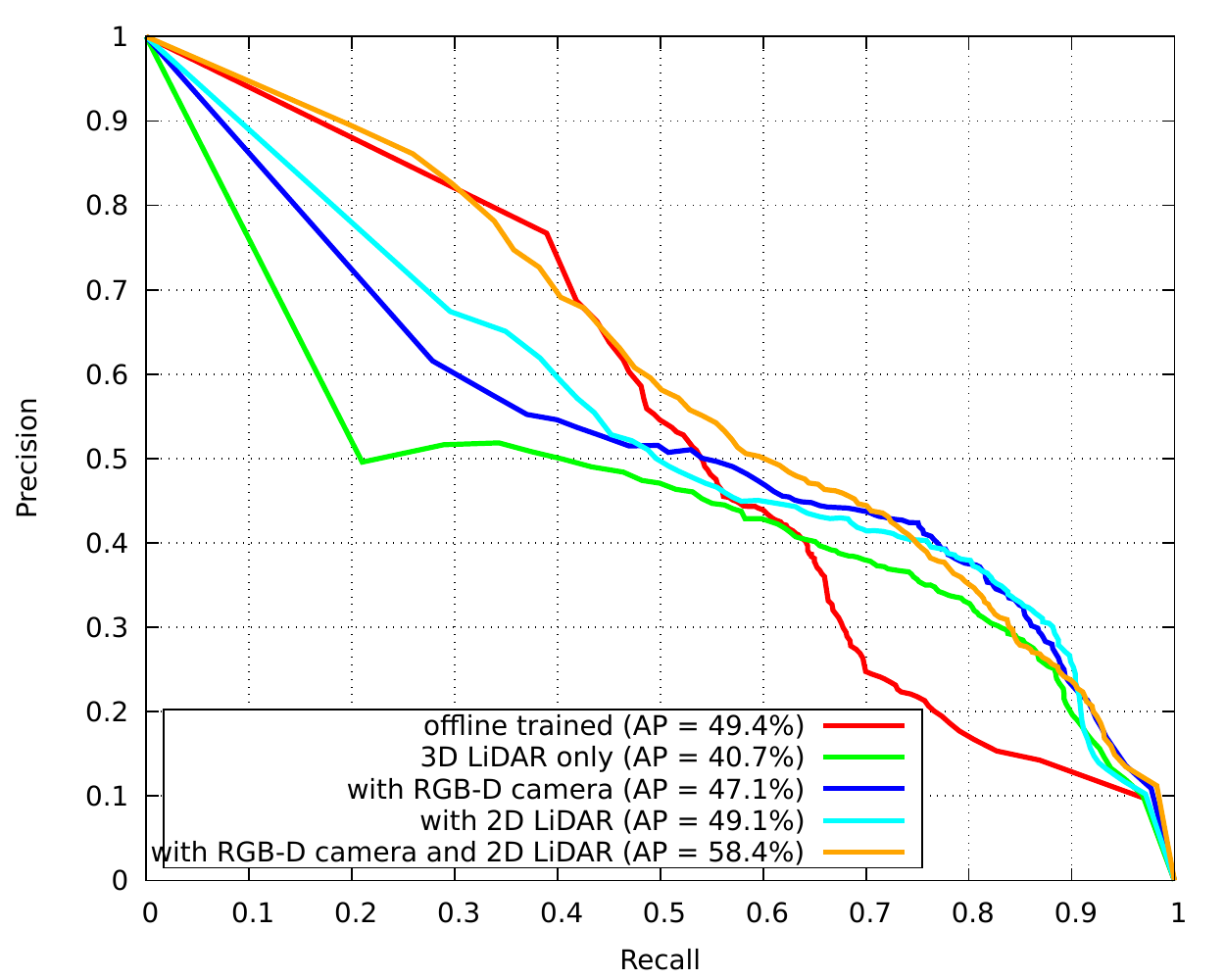}
  \caption{Performance evaluation of human detection. Best viewed in color.}
  \label{fig:classification_results}
\end{figure}

We additionally evaluate the Precision-Recall of the offline and final (i.e. after the 7th iteration) online trained classifiers.
The experimental results are shown in Fig.~\ref{fig:classification_results}.
Once again, the combination of RGB-D camera and 2D~LiDAR achieves the best performance.
Differently from the ``3D LiDAR only'' case, which learned only from moving people, the solution ``with RGB-D camera and 2D LiDAR'' was able to learn from moving, standing and sitting people, which greatly improved the human classification performance.
It is also worth pointing out that, despite showing a relatively high precision, the true positive rate (recall)  of the offline trained classifier is generally lower.
This is due to a lack of long-distance samples in the offline training set, which are difficult to label by a human annotator.

\section{CONCLUSION}\label{sec:conclusions}

In this paper, we presented a framework for online transfer learning, applied to 3D~LiDAR-based human classification, taking advantage of multisensor-based tracking.
The framework, which relies on the computation of human trajectory probabilities, enables a robot to learn a new human classifier over time with the help of existing human detectors.
To this end, we proposed a semi-supervised learning method, which fuses both model-based (labeled) and model-free (unlabeled) detections from different sensors.
A very promising feature of the proposed solution is that the new human classifier can be learned directly from the deployment environment, thus removing the dependence on pre-annotated data.
The experimental results, based on a real-world dataset, demonstrated the efficiency of our system.

The proposed framework has been fully implemented into ROS with a high level of modularity.
The software and the dataset are publicly available to the research community, with the intention to perform objective and systematic comparisons between the recognition capabilities of different robots.
Moreover, our framework is easy to extend to other sensors and moving objects, such as cars, bicycles and animals.

Despite these encouraging results, there are several aspects which could be improved.
For example, the AP of the online learned classifier is still relatively low, due to the complexity of the environment recorded in our dataset.
This can be further improved by using a more advanced model for negative sample generation.
In addition, it remains to be verified how a new human detector, based on the online trained classifier, will affect the stability of the system and its tracking performance.

\section*{ACKNOWLEDGMENT}

This work has received funding from the European Union's Horizon 2020 research and innovation programme under grant agreement No. 645376 (FLOBOT) and No. 732737 (ILIAD).

\bibliographystyle{IEEEtran}
\bibliography{references}

\end{document}